\pdfoutput=1
\documentclass[11pt]{article}
\usepackage{acl}
\usepackage{booktabs}
\usepackage{times}
\usepackage{latexsym}
\usepackage{amsmath,amsfonts,bm}
\usepackage[T1]{fontenc}
\usepackage[utf8]{inputenc}
\usepackage{xcolor}
\usepackage{pifont}
\usepackage{booktabs}
\usepackage{colortbl}
\usepackage{verbatim}
\usepackage{microtype}
\usepackage{multirow}
\usepackage{inconsolata}

\usepackage{graphicx}
\definecolor{text-red}{RGB}{194, 31, 48}
\definecolor{text-blue}{RGB}{36, 134, 185}
\definecolor{text-green}{RGB}{65, 174, 60}

\definecolor{table-green}{RGB}{173, 216, 200}
\definecolor{table-blue}{RGB}{173, 216, 230}
\title{\textsc{FastCuRL}: Curriculum Reinforcement Learning with Stage-wise \\Context Scaling for Efficient Training R1-like Reasoning Models}

\author{Mingyang Song, Mao Zheng, Zheng Li, Wenjie Yang, Xuan Luo, Yue Pan, Feng Zhang\\
	Tencent Hunyuan\\
	\texttt{nickmysong@tencent.com}\\
}
\begin{document}
\maketitle
\begin{abstract}
%Improving the training efficiency remains one of the biggest challenges in large-scale reinforcement learning (RL). To this end, this work investigates how \textbf{\textit{context length}} and \textbf{\textit{the complexity of training data}} influence \textit{the RL training process of R1-distilled reasoning models}. 

Improving training efficiency continues to be one of the primary challenges in large-scale Reinforcement Learning (RL). In this paper, we investigate how \textbf{\textit{context length}} and \textbf{\textit{the complexity of training data}} influence \textit{the RL scaling training process of R1-distilled reasoning models, e.g., DeepSeek-R1-Distill-Qwen-1.5B}.
\textit{Our experimental results reveal that:} {\color{text-green}\textit{(1) simply controlling the context length and selecting the training data based on the input prompt length can effectively improve the training efficiency of RL scaling, achieving better performance with more concise CoT;}} {\color{text-blue}\textit{(2) properly scaling the context length helps mitigate entropy collapse;}} {\color{text-red}\textit{and (3) carefully choosing the context length facilitates achieving efficient LLM training and reasoning}}. Inspired by these insights, we propose \textbf{\textsc{FastCuRL}}, a curriculum RL framework with stage-wise context scaling to achieve efficient LLM training and reasoning. Extensive experimental results demonstrate that \textbf{\textsc{FastCuRL}-1.5B-V3} significantly outperforms state-of-the-art reasoning models on five competition-level benchmarks and achieves 49.6\% accuracy on AIME 2024. Furthermore, \textbf{\textsc{FastCuRL}-1.5B-Preview} surpasses DeepScaleR-1.5B-Preview on five benchmarks while only using a single node with 8 GPUs and a total of 50\% of training steps, as shown in Figure~\ref{preview}.
The code, training data, and models have been publicly released\footnote{\url{https://github.com/nick7nlp/FastCuRL}}.

\end{abstract}

\section{Introduction}\label{introduction}
Large Language Models (LLMs) have emerged as immensely potent AI instruments, showcasing extraordinary proficiency in comprehending natural language and executing downstream tasks~\cite{llm_survey_2023, llm_survey_2024, chen2025reasoningerasurveylong}. Lately, test-time scaling~\cite{tts,muennighoff2025s1} has demonstrated a robust correlation between extending the generation length of Chain-of-Thought (CoT)~\cite{cot} and improving the reasoning capabilities of LLMs.

\begin{figure}[t!]
	\begin{center}
		\includegraphics[scale=0.317]{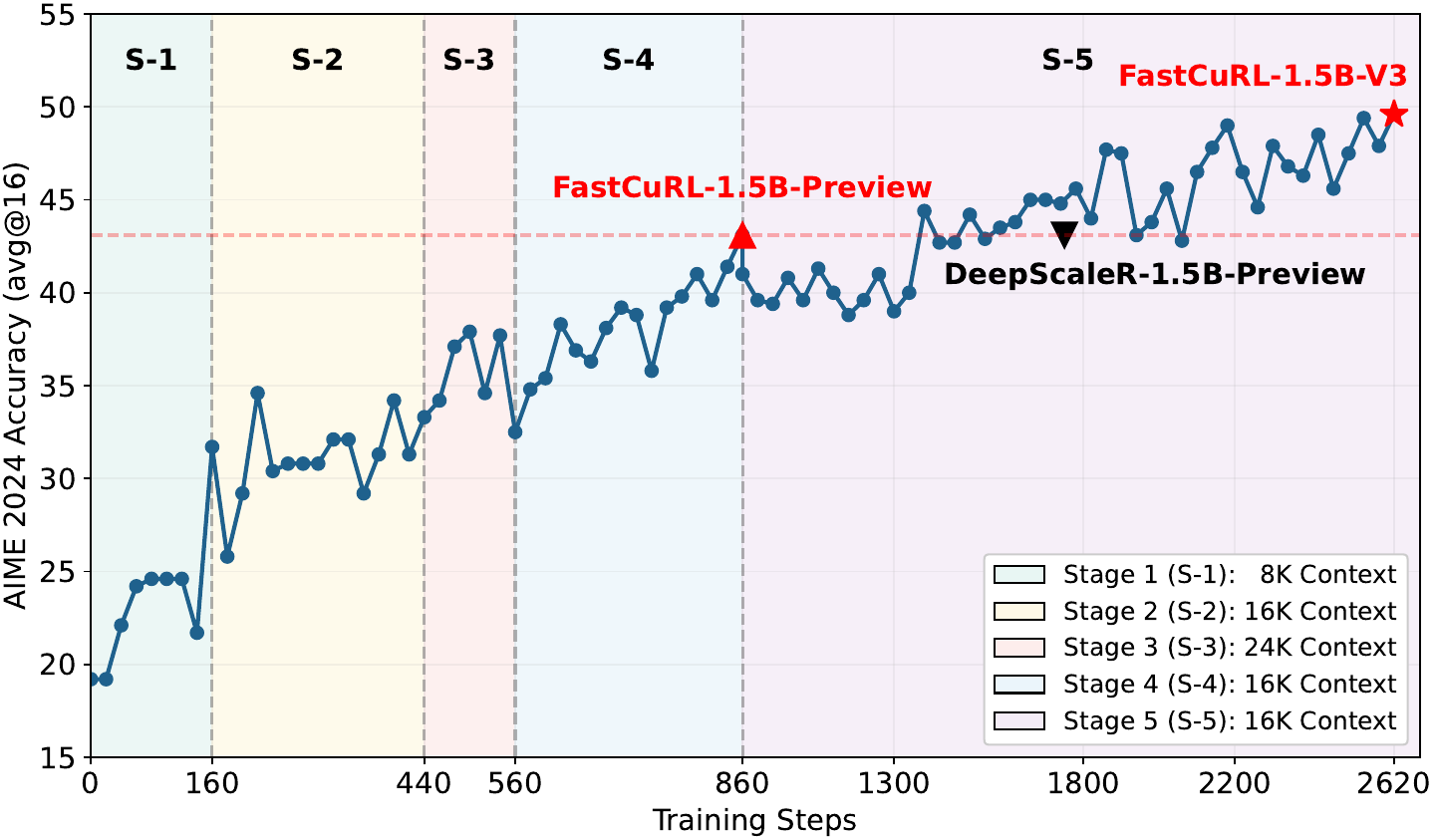}
	\end{center}
	\caption{FastCuRL’s accuracy on AIME 2024 as training progresses across five training stages. Specifically, S-5 indicates Stage 5 in the training process.}\label{preview}
\end{figure}

%A primary finding from recent breakthroughs, exemplified by DeepSeek-R1~\cite{ds-r1}, reveals a scaling phenomenon in the training process of Reinforcement Learning (RL).
%Inspired by these findings, training LLMs through scaling RL has recently emerged as a promising paradigm for addressing complex reasoning tasks and many valuable research endeavours~\cite{openr1,OpenReasonerZero2025,zeng2025simplerl,drgrpo} have emerged to explore and replicate reasoning models akin to DeepSeek-R1 (for example, starting from R1-distilled or pre-trained models) by extending the generation length of CoT. 
A primary finding from recent breakthroughs, exemplified by DeepSeek-R1 \cite{ds-r1}, have revealed a notable scaling phenomenon in the training process of Reinforcement Learning (RL). Inspired by these findings, scaling RL for training LLMs has emerged as a promising paradigm for tackling complex reasoning tasks. This has sparked numerous valuable research efforts aimed at exploring and replicating reasoning models similar to DeepSeek-R1~\cite{openr1,OpenReasonerZero2025,zeng2025simplerl,drgrpo}. These initiatives typically begin with R1-distilled or pre-trained models and extend the generation length of CoT reasoning to enhance performance.

However, excessively long CoT responses significantly increase computational overhead during model training and deployment. Moreover, recent studies~\cite{yeo2025demystifyinglongchainofthoughtreasoning,wu2025lessunderstandingchainofthoughtlength,kimi1.5} have identified an inherent overthinking phenomenon in reasoning models, which includes irrelevant details and repetitive thinking patterns. This kind of information leads to inefficient use of computational resources and undermines reasoning accuracy, which causes models to stray from valid logical pathways, resulting in incorrect answers.
To this end, recent studies~\cite{kimi1.5, deepscaler2025, drgrpo, song2025walkrunconcisellm} focus on efficient reasoning for optimizing the model to generate more concise CoT outputs. Among them, DeepScaleR~\cite{deepscaler2025} propose to iteratively increase the context length from 8K to 24K to train the \textsc{DeepSeek-R1-Distill-Qwen}-1.5B model toward more concise reasoning, outperforming OpenAI's o1-preview~\cite{openai2024gpt4}.
By observing the training logs\footnote{\url{https://github.com/agentica-project/rllm}} of DeepScaleR in Figure~\ref{deepscaler}, we find two issues:
\begin{figure}[t!]
	\begin{center}
		\includegraphics[scale=0.26]{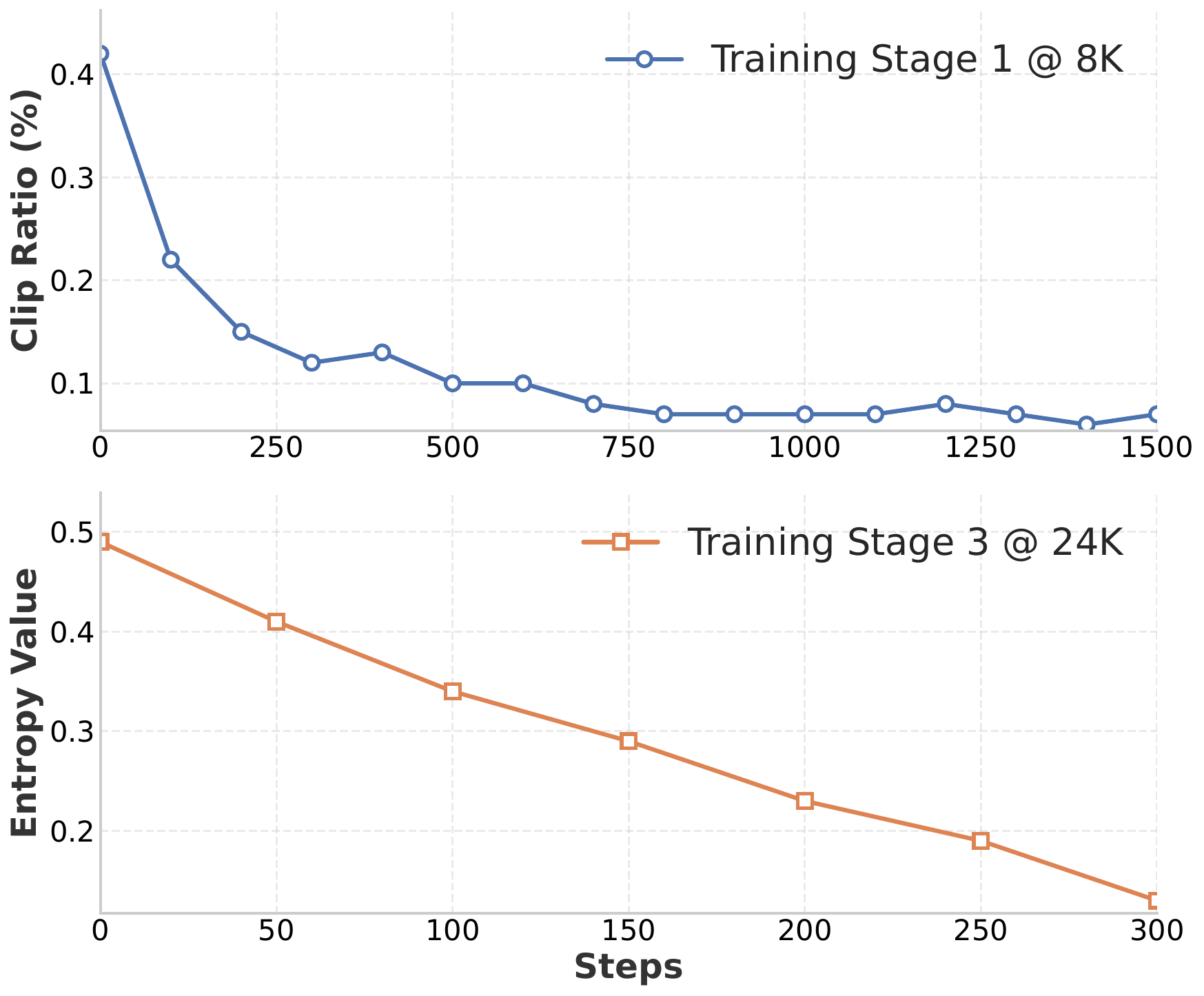}
	\end{center}
	\caption{Partial training curves of DeepScaleR. Data points are smoothed using a running average (window size = 10) and plotted at uniform intervals.}\label{deepscaler}
\end{figure}
\begin{itemize}
	\item When the context length is 8K, about 42\% of the model's outputs are clipped, which reduces the training efficiency.
	\item When training in the third stage with a context length of 24K, the model's entropy collapses\footnote{Entropy reflects the exploration capability of LLMs during RL training. Typically, a rapid decrease in entropy might lead to premature convergence, preventing the model from achieving the expected performance.}, reducing its exploration capabilities.
\end{itemize}

The prior work and the aforementioned issues naturally motivate two research questions:
\begin{itemize}
	\item \textit{\textbf{Question 1}}: \textit{Does simultaneously controlling the model's context length and the complexity of the training dataset help the training process of R1-like reasoning models?}
	\item \textit{\textbf{Question 2}}: \textit{What impact does setting different context lengths have on the RL training process of R1-like reasoning models?}
\end{itemize}

To this end, in this paper, we investigate how the model’s context length and the complexity of the training dataset influence the training process of R1-like reasoning models.
Motivated by our observations, we propose \textbf{\textsc{FastCuRL}}, \textit{a simple yet efficient} \textbf{Cu}rriculum \textbf{R}einforcement \textbf{L}earning framework with a stage-wise context scaling strategy to improve the RL training efficiency and achieve concise CoT. 
Specifically, the proposed method alternates between CoT compression (long-to-short) and extension (short-to-long). We first compress CoT reasoning outputs, then extend them, and repeat this compress–extend cycle. The main goal of our method is to enhance CoT quality progressively. We hypothesize that this dynamic interplay fosters higher-quality reasoning by iteratively refining essential logical dependencies while pruning redundant or spurious steps.
Meanwhile, during the iterative process, we control the complexity of the training samples, thereby ensuring training efficiency.
Extensive experimental results demonstrate that our model \textbf{\textsc{FastCuRL}-1.5B-V3} outperforms recent state-of-the-art reasoning baselines across five competition-level benchmarks, AIME 2024, AMC 2023, MATH 500, Minerva Math, and OlympiadBench. Furthermore, our model \textbf{\textsc{FastCuRL}-1.5B-Preview} surpasses DeepScaleR-1.5B-Preview on five competition-level benchmarks and only uses 50\% training steps on a single node with 8 GPUs. We hope the findings presented in this paper, the models we have released, and the open-sourced code will benefit future research.

\section{Methodology}
Our method for achieving efficient LLM reasoning integrates three main components: (1) a resource-efficient RL algorithm, (2) a complexity-aware, mathematics-focused dataset, and (3) a stage-wise context scaling approach. This section provides a detailed description of the first two components.
\begin{figure}[t!]
	\begin{center}
		\includegraphics[scale=0.25]{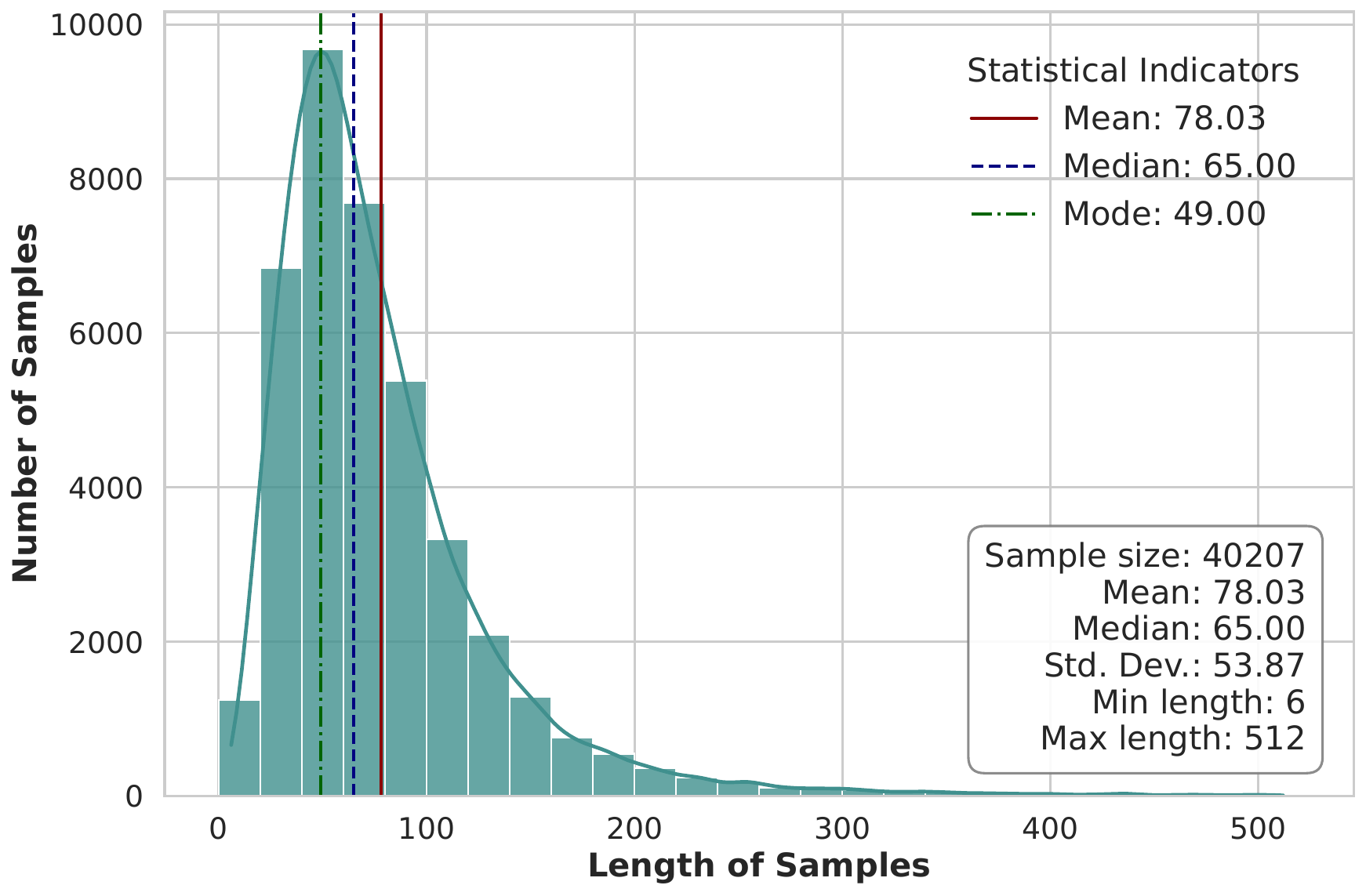}
	\end{center}
	\caption{Prompt length distribution. The prompt length is computed based on the number of tokens.}\label{length_distribution}
\end{figure}
\subsection{Reinforcement Learning Algorithm}
To train our model efficiently, we adopt the Group Relative Policy Optimization (GRPO)~\cite{deepseek-math}, which is utilized in~\citet{ds-r1}. GRPO eliminates the necessity of maintaining a critic model, which is usually comparable in size to the policy model, by estimating baseline scores directly from group-level scores, significantly lowering the computational overhead. For each problem $q$, GRPO directly samples a group of $\mathrm{G}$ responses $\{o_1, o_2, ..., o_G\}$ from the old policy $\pi_{\theta_{\mathrm{old}}}$ and optimizes the trained policy $\pi_{\theta}$ by maximizing the following objective:
\begin{equation}
	\small
	\begin{aligned}
		\mathcal{J}_{\mathrm{GRPO}}(\theta) = &\ \mathbb{E}_{[q \sim P(Q), \{o_i\}_{i=1}^{\mathrm{G}} \sim \pi_{\theta_{\mathrm{old}}}(\cdot|q)]} \\
		&\frac{1}{\mathrm{G}} \sum_{i=1}^{\mathrm{G}}\Bigg(\text{min}\Bigg(\frac{\pi_\theta(o_{i}|q)}{\pi_{\theta_{\mathrm{old}}}(o_{i}|q)}\mathrm{A}_i, \\
		&\mathrm{clip}\Bigg(\frac{\pi_\theta(o_i|q)}{\pi_{\theta_{\mathrm{old}}}(o_i|q)}, 1-\varepsilon, 1+\varepsilon\Bigg)\mathrm{A}_i\Bigg) \\&- \beta\mathbb{D}_{\text{KL}}[\pi_\theta||\pi_{\mathrm{ref}}]\Bigg),
	\end{aligned}
\end{equation}
where $\beta$ denotes the coefficient of the KL penalty and the advantage $\mathrm{A}_i$ is computed from a group of rewards $\{r_1, r_2, ..., r_{\mathrm{G}}\}$:
\begin{equation}
	\small
	\mathrm{A}_i = \frac{r_i-\mathrm{mean}(\{r_1, r_2, ..., r_{\textit{G}}\})}{\mathrm{std}(\{r_1, r_2, ..., r_{\mathrm{G}}\})}.
\end{equation}
To enhance the exploration capabilities of the policy model, we introduce an entropy bonus term,
\begin{equation}
	\small
	\begin{aligned}
		\mathcal{J}_{\mathrm{GRPO}^{\dagger}}(\theta) = &\ \mathbb{E}_{[q \sim P(Q), \{o_i\}_{i=1}^{\mathrm{G}} \sim \pi_{\theta_{\mathrm{old}}}(\cdot|q)]} \\
		&\frac{1}{\mathrm{G}} \sum_{i=1}^{\mathrm{G}}\Bigg(\text{min}\Bigg(\frac{\pi_\theta(o_{i}|q)}{\pi_{\theta_{\mathrm{old}}}(o_{i}|q)}\mathrm{A}_i, \\
		&\mathrm{clip}\Bigg(\frac{\pi_\theta(o_i|q)}{\pi_{\theta_{\mathrm{old}}}(o_i|q)}, 1-\varepsilon, 1+\varepsilon\Bigg)\mathrm{A}_i\Bigg) \\&- \beta\mathbb{D}_{\text{KL}}[\pi_\theta||\pi_{\mathrm{ref}}]+\underline{\alpha \mathbb{H}(\pi_\theta(o_i|q))}\Bigg),
	\end{aligned}
\end{equation}
where $\mathbb{H}(\pi_\theta(o_i|q))$ indicates the entropy term and $\alpha$ denotes the entropy coefficient.

Similar to the prior work~\cite{ds-r1, deepscaler2025}, we leverage a rule-based reward model composed of two distinct criteria designed to balance answer correctness and clarity of structure without relying on an LLM-based reward model. To evaluate correctness objectively, we require the trained model to present its final answer enclosed within a \verb|\|\textbf{boxed\{\}} format, assigning a binary score of 1 for correct answers and 0 for incorrect ones. To encourage structural clarity, the model must explicitly encapsulate its reasoning within tags, with compliance being rewarded positively.

\subsection{Complexity-Aware Data Selection}
To ensure a fair comparison, we directly employ the dataset from DeepScaleR as the training data. The DeepScaleR dataset~\cite{deepscaler2025} consists of 40,315 unique mathematics-specific problem-answer pairs collected from AIME (1984-2023), AMC (prior to 2023), Omni-MATH, and the Still dataset~\cite{matharena, OmniMATH, still_dataset}. We perform a simple filtering of the original dataset based on prompt length, resulting in 40,207 data samples. The statistics of the filtered dataset are shown in Figure~\ref{length_distribution}.

\begin{figure}[t!]
	\begin{center}
		\includegraphics[scale=0.265]{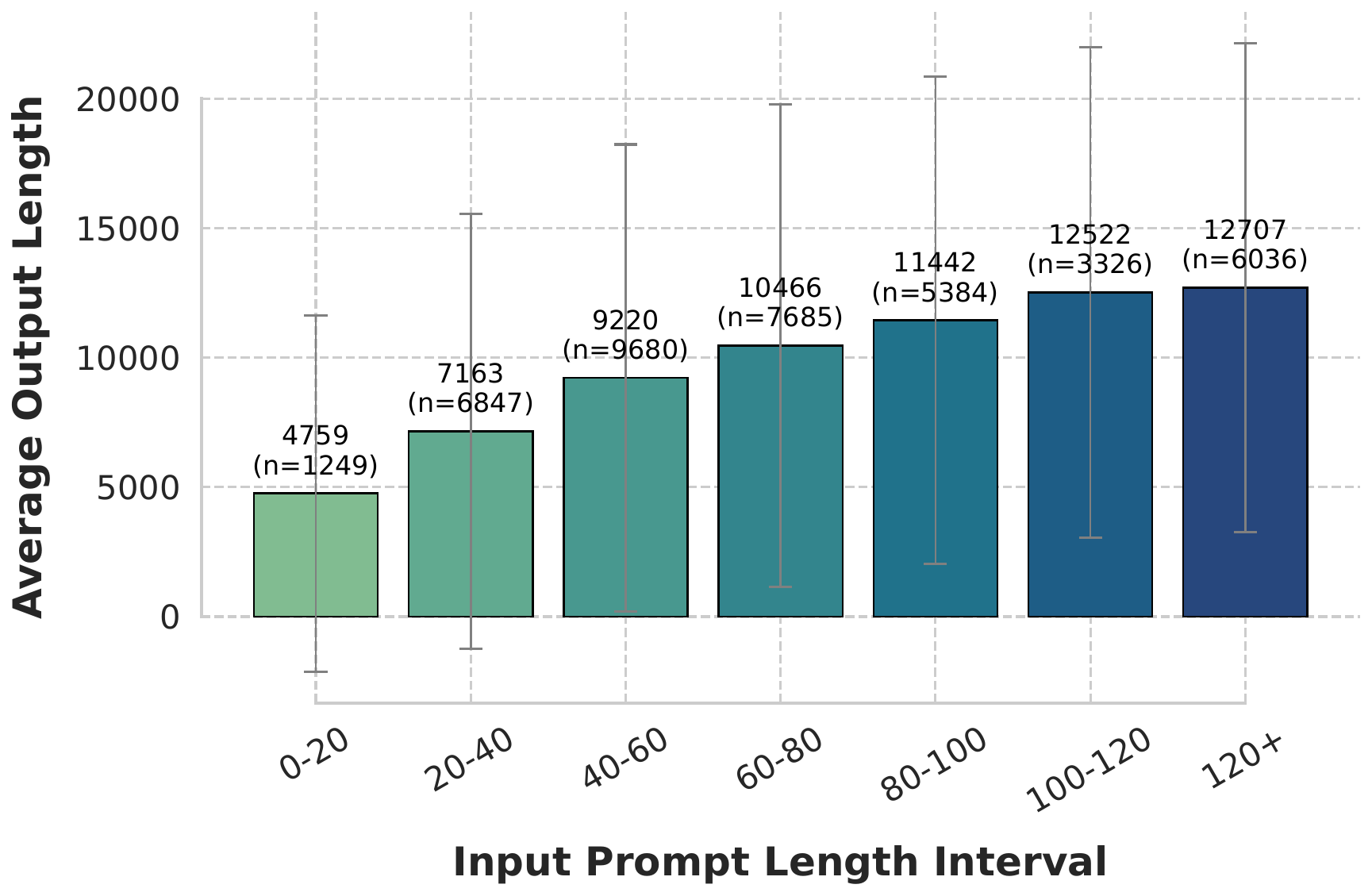}
	\end{center}
	\caption{Distribution of output length across different input prompt length intervals in the training dataset. This distribution illustrates the average output length (in tokens) generated by \textsc{DeepSeek-R1-Distill-Qwen}-1.5B, with error bars indicating standard deviation. Each bar is annotated with the mean output length and sample size ($n$) for the corresponding input length interval.}\label{input_output_relation}
\end{figure}
\begin{table}[t!]
	\small
	\begin{center}
		\renewcommand\tabcolsep{7pt}
		\renewcommand\arraystretch{1.2}
		\begin{tabular}{cccc}
			\toprule
			\multicolumn{1}{c}{\textbf{Data}}  &\multicolumn{1}{c}{\textbf{Prompt Len.}}&\multicolumn{1}{c}{\textbf{Output Len.}} &\textbf{\#Samples}\\
			\midrule
			$\textbf{L}$1 &48.26&8805.78&25168\\
			$\textbf{L}$2  &78.03&10063.44&40207\\
			$\textbf{L}$3  &132.22&12168.16&15039\\
			\bottomrule
		\end{tabular}
	\end{center}
	\caption{Statistics of $\textbf{L}$1, $\textbf{L}$2, $\textbf{L}$3. The prompt and output length is computed based on the number of tokens.}\label{three}
\end{table}

\begin{comment}
\begin{table}[t!]
	\centering
	\scriptsize
	\renewcommand\arraystretch{1.3}
	\begin{tabular}{@{}p{0.98\linewidth}@{}}
		\hline\hline
		\textbf{Example Problem} \textit{(Output Length=74706 characters)}: Ashley, Betty, Carlos, Dick, and Elgin went shopping. Each had a whole number of dollars to spend, and together they had $56$ dollars. The absolute difference between the amounts Ashley and Betty had to spend was $19$ dollars. The absolute difference between the amounts Betty and Carlos had was $7$ dollars, between Carlos and Dick was $5$ dollars, between Dick and Elgin was $4$ dollars, and between Elgin and Ashley was $11$ dollars. How many dollars did Elgin have?  \\\hline\hline
	\end{tabular}
	\caption{Example problem.}
	\label{description}
\end{table}
\end{comment}

\begin{figure*}[t!]
	\begin{center}
		\includegraphics[scale=0.43]{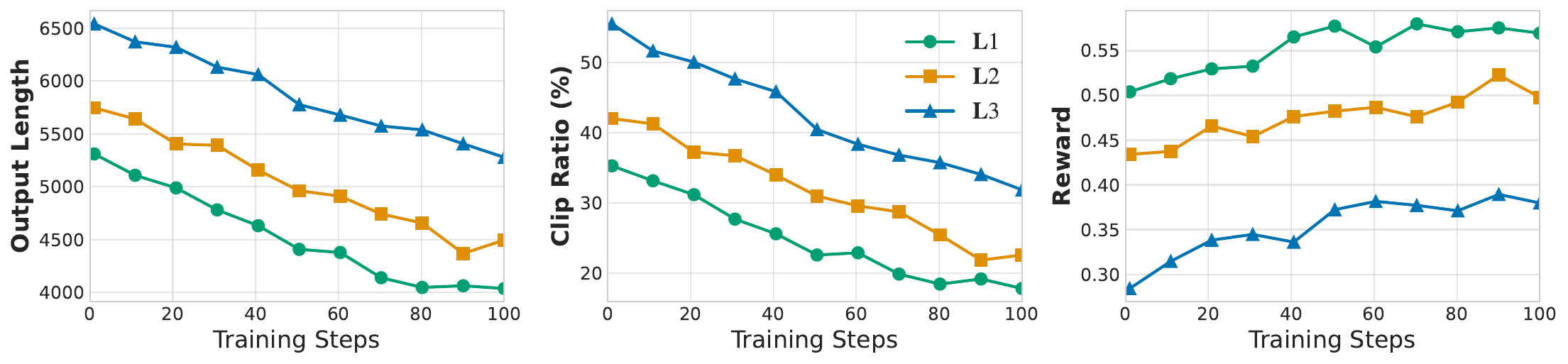}
	\end{center}
	\caption{The average clip ratio, output length, and reward in RL training with 8K context length on $\mathbf{L}$1, $\mathbf{L}$2, and $\mathbf{L}$3 datasets. Data points are smoothed using a running average (window size = 10) and plotted at uniform intervals.}\label{compare}
\end{figure*}
As illustrated in Figure~\ref{deepscaler}, during the first training stage, over 42\% of training samples are clipped at the beginning of the training steps due to exceeding the maximum response length. By observing and analyzing the clipped responses, we find that they mainly correspond to two types of problems. The first type pertains to challenging problems requiring long CoT responses to solve. The second involves questions laden with numerous conditions, prompting the model to verify each condition repeatedly during problem-solving. This repetitive verification may result in redundant thinking patterns, ultimately causing the reasoning responses to be unduly long. Both situations may impact the model training efficiency during the 8K context. 

After observing the above phenomenon, we utilize \textsc{DeepSeek-R1-Distill-Qwen}-1.5B to infer all the training data of DeepScaleR to obtain responses and analyze the response lengths, as shown in Figure~\ref{input_output_relation}. Specifically, the given figure examines the relationship between input length and output length. Interestingly, we find a correlation between the two-that is, the longer the input, the longer the corresponding output. Based on this observation, we assume a hypothesis that for complex reasoning tasks, there exists a relationship between the complexity of the problem prompt and the length of the output response generated by the model when solving it. Generally, the more complex the problem, the longer the output the model needs to produce to arrive at a solution. Based on this hypothesis, we directly divide the original training dataset (referred to as $\textbf{L}$2) into two training data subsets based on the average input prompt length: one representing a short CoT reasoning dataset (designated as $\textbf{L}$1) and the other constituting a long CoT reasoning dataset (labeled as $\textbf{L}$3). Finally, the average input length of each dataset as shown in Table~\ref{three}.

Next, we conduct experiments and analyses on these three datasets under different context lengths to observe and investigate the two questions raised in the prior section. It is important to note that this paper focuses on low-resource scenarios. Therefore, during training, when using different datasets at each stage, we train for only one epoch and utilize a single node with 8 GPUs.

\section{Experiments}
We designed a set of experiments, constrained by available computational resources, to investigate our central research question (Section~\ref{introduction}): \textit{how do model context length and training data complexity affect the RL training of R1-like reasoning models?} By analyzing the training behavior of small LLMs, we aim to extract practical insights. The objective is to move beyond simply providing empirical evidence of performance gains and to offer clear, actionable guidance for both academic and industrial implementation.
\subsection{Experimental Setup}
In this work, we choose a 1.5B parameter model \textsc{DeepSeek-R1-Distill-Qwen}-1.5B~\cite{ds-r1} as the base model. We utilize the AdamW optimizer with a constant learning rate of $1\times10^{-6}$ for optimization. For rollout, we set the temperature to 0.6 and sample 16 responses per prompt. We do not utilize a system prompt; instead, we add "Let's think step by step and output the final answer within \verb|\|\textbf{boxed\{\}}." at the end of each problem. Detailed parameters are shown in Figure~\ref{exp}.
\begin{table*}[t!]
	\scriptsize
	\begin{center}
		\renewcommand\tabcolsep{3.3pt}
		\renewcommand\arraystretch{1.2}
		\begin{tabular}{lllllllllc}
			\toprule
			\multirow{1}{*}{\textsc{\textbf{Experiments}}}&\multirow{1}{*}{\textsc{\textbf{Training}}}&\multicolumn{2}{l}{\textsc{\textbf{Context Length}}}&\multirow{2}{*}{\textsc{\textbf{Training Data}}}&\multirow{2}{*}{\textsc{\textbf{Batch Size}}}&\multirow{2}{*}{\textsc{\textbf{Rollout}}}&\multirow{1}{*}{$\alpha$}&\multirow{1}{*}{$\beta$}&\multirow{2}{*}{\textsc{\textbf{Avg.}}}\\
			\multirow{1}{*}{\textsc{\textbf{(Stages)}}}&\multirow{1}{*}{\textsc{\textbf{Steps}}}&\textsc{\textbf{Input}}&\textsc{\textbf{Output (K)}}&&&&\textsc{\textbf{(for Entropy)}}&\textsc{\textbf{(for KL)}}&\\
			\midrule
			%\textsc{DeepScaleR} &3&1K&8, 16, 24&$\mathbf{L}$2, $\mathbf{L}$2, $\mathbf{L}$2&128, 128, 128&8, 16, 16& 0.001 & 0.001& 0.569\\
			%\midrule
			\textsc{Exp}-1 (3)&3&1K&8, 16, 24&$\mathbf{L}$1, $\mathbf{L}$2, $\mathbf{L}$3&128, 64, 64&8, 8, 8& 0.001 & 0.001& 0.550\\
			\textsc{Exp}-2 (3)&3&1K&8, 16, 24&$\mathbf{L}$1, $\mathbf{L}$3, $\mathbf{L}$2&128, 64, 64&8, 8, 8&0.001&0.001&0.540\\
			\textsc{Exp}-3 (3)&3&1K&8, 16, 24&$\mathbf{L}$1, $\mathbf{L}$2, $\mathbf{L}$2&128, 64, 64&8, 8, 8&0.001&0.001&0.552\\
			
			\midrule
			
			\textsc{Exp}-4 (4)&4&1K&8, 16, 24, \textbf{32}&$\mathbf{L}$1, $\mathbf{L}$2, $\mathbf{L}$3, $\mathbf{L}$2&128, 64, 64, 64 &8, 8, 8, 16&0.001& 0.001&0.566\\
			\textsc{Exp}-5 (4)&4&1K&8, 16, 24, \textbf{24}&$\mathbf{L}$1, $\mathbf{L}$2, $\mathbf{L}$3, $\mathbf{L}$2&128, 64, 64, 64 &8, 8, 8, 16&0.001&0.001&0.565\\
			
			\midrule
			
			\rowcolor{table-blue}\textsc{Exp}-6 (4)&4&1K&8, 16, 24, \textbf{16}&$\mathbf{L}$1, $\mathbf{L}$2, $\mathbf{L}$3, $\mathbf{L}$2&128, 64, 64, 64 &8, 8, 8, 16&0.001&0.001&0.575\\
			
			\midrule
			
			\textsc{Exp}-7 (5)&5&1K&8, 16, 24, 16, \textbf{8}&$\mathbf{L}$1, $\mathbf{L}$2, $\mathbf{L}$3, $\mathbf{L}$2, $\mathbf{L}$2&128, 64, 64, 64, 64 &8, 8, 8, 16, 16&0.001&0.001&0.535\\
			\textsc{Exp}-8 (5)&5&1K&8, 16, 24, 16, \textbf{16}&$\mathbf{L}$1, $\mathbf{L}$2, $\mathbf{L}$3, $\mathbf{L}$2, $\mathbf{L}$2&128, 64, 64, 64, 64 &8, 8, 8, 16, 16&0.001&0.001&0.567\\
			\textsc{Exp}-9 (5)&5&1K&8, 16, 24, 16, \textbf{24}&$\mathbf{L}$1, $\mathbf{L}$2, $\mathbf{L}$3, $\mathbf{L}$2, $\mathbf{L}$2&128, 64, 64, 64, 64 &8, 8, 8, 16, 16&0.001& 0.001&0.556\\
			
			\midrule
			
			\rowcolor{table-green}\textsc{Exp}-10 (5)&5&1K&8, 16, 24, 16, \textbf{16}&$\mathbf{L}$1, $\mathbf{L}$2, $\mathbf{L}$3, $\mathbf{L}$2, $\mathbf{L}$2&128, 64, 64, 64, 64 &8, 8, 8, 16, 16&\textbf{0.000001}&\textbf{0.000}&0.600\\
			
			\midrule
			
			\rowcolor{table-green}\textsc{Exp}-11 (5)&5&1K&8, 16, 24, 16, \textbf{16}&$\mathbf{L}$1, $\mathbf{L}$2, $\mathbf{L}$3, $\mathbf{L}$2, $\mathbf{L}$2&128, 64, 64, 64, 64 &8, 8, 8, 16, 16&\textbf{0.000}&\textbf{0.000}&0.616\\
			\bottomrule
		\end{tabular}
	\end{center}
	\caption{Experimental setups combining different context lengths and data complexities.}\label{exp}
\end{table*}
\begin{table*}[t!]
	\small
	\begin{center}
		\renewcommand\tabcolsep{2pt}
		\renewcommand\arraystretch{1.}
		\begin{tabular}{lcccccc}
			\toprule
			\multicolumn{1}{l}{\textbf{Model}}  &\multicolumn{1}{c}{\textbf{MATH 500}} &\multicolumn{1}{c}{\textbf{AIME 2024}}  &\multicolumn{1}{c}{\textbf{AMC 2023}}  &\multicolumn{1}{c}{\textbf{Minerva Math}}&\multicolumn{1}{c}{\textbf{OlympiadBench}} &\multicolumn{1}{c}{\textbf{Avg. Score}}  \\
			\midrule
			\textsc{Qwen2.5-Math}-7B-Instruct &79.8&13.3&50.6&34.6&40.7&43.8 \\
			\textsc{rStar-Math}-7B &78.4&26.7&47.5&-&47.1&- \\
			\textsc{Eurus}-2-7B-PRIME &79.2&26.7&57.8&38.6&42.1&48.9 \\
			\textsc{Qwen}2.5-7B-SimpleRL &82.4&26.7&62.5& 39.7 &43.3&50.9  \\\midrule\midrule
			\textsc{DeepSeek-R1-Distill-Qwen}-1.5B  &82.8&28.8&62.9&26.5&43.3&48.9 \\
			\textsc{Still}-3-1.5B-Preview &84.4&32.5&66.7&29.0&45.4&51.6 \\
			\textsc{DeepScaleR}-1.5B-Preview &87.8&43.1&73.6&30.2&50.0&56.9\\\midrule
			\textbf{\textsc{FastCuRL}-1.5B-Preview} & {88.0} & {43.1} & {74.2} & {31.6} & {50.4} & {57.5}\\
			\textbf{\textsc{FastCuRL}-1.5B-V2} & {89.3} & {47.5} & {77.0} & {32.8} & {53.3} & {60.0}\\
			\textbf{\textsc{FastCuRL}-1.5B-V3} & \textbf{90.5} & \textbf{49.6} & \textbf{78.5} & \textbf{34.7} & \textbf{54.5} & \textbf{61.6}\\
			\bottomrule
		\end{tabular}
	\end{center}
	\caption{\textsc{Pass@1} performance across competition-level mathematical benchmarks.}\label{op}
\end{table*}
\subsection{Baselines}
In this paper, we conduct evaluations against 1.5B and 7B parameter language models, which includes \textsc{DeepSeek-R1-Distill-Qwen}-1.5B~\cite{ds-r1}, \textsc{Qwen2.5-Math}-7B-Instruct~\cite{Qwen2.5-Math}, DeepScaleR-1.5B-Preview~\cite{deepscaler2025}, \textsc{Qwen}2.5-7B-SimpleRL~\cite{zeng2025simplerl}, \textsc{rStar-Math}-7B~\cite{rStar-Math}, \textsc{Still}-3-1.5B-Preview~\cite{Slow_Thinking_with_LLMs_3_Preview}, and \textsc{Eurus}-2-7B-PRIME~\cite{Eurus-2-7B-PRIME}.
\subsection{Benchmarks}
To comprehensively evaluate the performance, we select five competition-level benchmark datasets: MATH 500~\cite{MATH}, AIME 2024\footnote{\url{https://huggingface.co/datasets/AI-MO/aimo-validation-aime}}, AMC 2023\footnote{\url{https://huggingface.co/datasets/AI-MO/aimo-validation-amc}}, Minerva Math~\cite{Minerva}, and OlympiadBench~\cite{OlympiadBench}.

\subsection{Evaluation Metric}
Following the prior work~\cite{ds-r1}, we set the maximum context length to 32,768 tokens and use \textsc{Pass@1}  as the evaluation metric. Specifically, we adopt \textbf{a sampling temperature of $\mathbf{0.6}$} and \textbf{a top-p value of $\mathbf{1.0}$} to generate $k$ responses for each question, typically $k=16$. Specifically, \textsc{Pass@1} is then calculated as:
\begin{equation}
	\textsc{Pass@1} = \frac{1}{k}\sum_{i=1}^{k}p_i,
\end{equation}
where $p_i$ is the correctness of the $i$-th response.

\subsection{Main Processes and Results}
In this section, we first validate the effectiveness of the complexity-aware data selection strategy. Then, we design a series of progressive experiments with varying context lengths and data complexities and analyze the experimental results.

\begin{table*}[t!]
	\small
	\begin{center}
		\renewcommand\tabcolsep{10pt}
		\renewcommand\arraystretch{1.05}
		\begin{tabular}{llcl}
			\toprule
			\multicolumn{1}{l}{Model}  &\multicolumn{1}{c}{Training Steps}&\multicolumn{1}{l}{Training Stages}  &\multicolumn{1}{l}{Number of GPUs Used in Each Stage}\\
			\midrule
			\textsc{DeepScaleR}-1.5B-Preview &$\sim1,750$&3&8, 16, 32 \\
			\textbf{\textsc{FastCuRL}-1.5B-Preview} (\textsc{Exp}-6)&$\sim860$&4&8, 8, 8, 8\\
			\textbf{\textsc{FastCuRL}-1.5B-V2} (\textsc{Exp}-10)&$\sim1,710$&5&8, 8, 8, 8, 8\\
			\textbf{\textsc{FastCuRL}-1.5B-V3} (\textsc{Exp}-11)&$\sim2,620$&5&8, 8, 8, 8, 8\\
			\bottomrule
		\end{tabular}
	\end{center}
	\caption{Training configurations and computational resources for different model variants. For fair comparison, training steps are normalized such that two steps with batch size 64 are equivalent to one step with batch size 128.}\label{td}
\end{table*}
\begin{figure*}[t!]
	\begin{center}
		\includegraphics[scale=0.355]{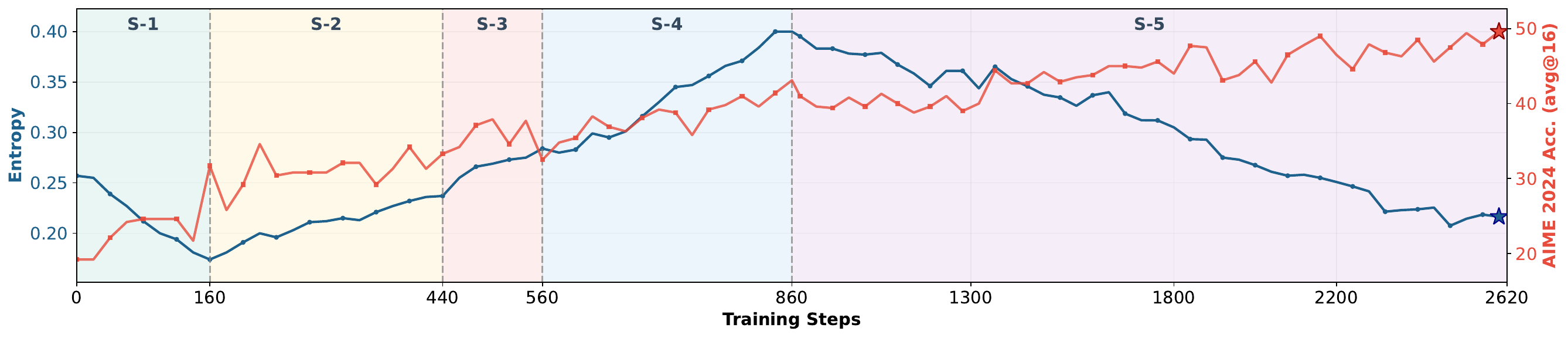}
	\end{center}
	%\caption{Entropy values of \textsc{FastCuRL}-1.5B-Preview and \textsc{FastCuRL}-1.5B-V3 across all RL training stages. Here, data points are smoothed using a running average (window size = 10) and plotted at uniform intervals.}\label{entropy_loss_exp6}
	\caption{Entropy values of our models across all RL training stages. Here, data points are smoothed using a running average (window size = 10) and plotted at uniform intervals.}\label{entropy_reward_5_stages}
\end{figure*}
\subsubsection{Main Results}
Table~\ref{op} presents a comprehensive comparative analysis of the \textsc{Pass@1} performance of our models against several state-of-the-art open-source baselines. The results unequivocally demonstrate the superiority of our proposed approach. Specifically, our final model, \textsc{FastCuRL}-1.5B-V3, achieves an average score of 61.6, significantly outperforming all baselines, including those with larger parameters. More specifically, \textsc{FastCuRL}-1.5B-V3 establishes new state-of-the-art results across all five competition-level benchmarks, scoring 90.5 on MATH 500, 49.6 on AIME 2024, 78.5 on AMC 2023, 34.7 on Minerva Math, and 54.5 on OlympiadBench. The consistent performance progression from \textsc{FastCuRL}-1.5B-Preview (57.5) to V3 (61.6) further validates the effectiveness and scalability of our training method.

In addition to overall performance, our models exhibit enhanced generalization capabilities. The initial \textsc{FastCuRL}-1.5B-Preview model already shows improvement over the \textsc{DeepScaleR}-1.5B-Preview on the AMC 2023 (74.2 vs. 73.6) and Minerva Math (31.6 vs. 30.2) test sets. This advantage is substantially amplified in our final model, \textsc{FastCuRL}-1.5B-V3, which widens the performance gap with scores of 78.5 and 34.7, respectively. From the perspective of training efficiency, these superior results are achieved with remarkable resource savings. Compared to the training regimen of the \textsc{DeepScaleR}-1.5B-Preview, our method utilizes only 50\% of the training steps. It requires just a single node with 8 GPUs, thereby reducing training costs by more than half, as shown in Table~\ref{td}. This demonstrates that our approach not only yields a more powerful model but also represents a more computationally efficient for developing advanced reasoning capabilities.

\subsubsection{Analyzing Entropy in RL Training}
Figure~\ref{entropy_reward_5_stages} presents the policy entropy and model performance across all training steps, segmented into five distinct stages (S-1 through S-5). The entropy trajectory exhibits three characteristic regimes: (i) gradual increase from 0.17 to 0.28 during initial exploration (S-1 to S-3), (ii) peak exploration at 0.40 during S-4, and (iii) substantial collapse to 0.22 in the final stage (S-5). 

A central challenge in RL scaling is managing the exploration-exploitation trade-off, where a standard failure mode is "entropy collapse." This phenomenon occurs when the policy rapidly becomes deterministic, leading to a sharp decrease in entropy and causing the policy to prematurely converge to a sub-optimal local optimum, thereby losing its ability to discover more effective strategies. However, as depicted in Figure~\ref{entropy_reward_5_stages}, our approach deviates significantly from this standard pattern. Instead of a premature collapse, the policy entropy demonstrates a prolonged and even increasing stage of exploration (S-2 to S-4), indicating that the policy actively maintains stochasticity. This sustained exploration is critical for navigating complex solution spaces and avoiding early stagnation.

By adjusting the reasoning context length and training data complexity, we effectively reinvigorate the policy's exploratory drive at key training intervals. The pronounced entropy peak in S-4 exemplifies this effect, where the agent is stimulated to explore novel behaviors rather than settling for its current policy. Crucially, this period of heightened exploration lays the groundwork for a more effective exploitation stage in S-5. As the policy finally converges and entropy decreases, the model's performance, measured by AIME 2024 Accuracy, continues to ascend to its peak. This demonstrates that our method successfully mitigates premature entropy collapse, enabling the model to achieve a significantly higher performance ceiling. 

\begin{table*}[t!]
	\small
	\begin{center}
		\renewcommand\tabcolsep{4pt}
		\renewcommand\arraystretch{1.3}
		\begin{tabular}{lcccccc}
			\toprule
			\multirow{2}{*}{Model} &\multicolumn{3}{c}{\# Average Output Length} &\multicolumn{3}{l}{\# Average Frequency of "Wait" and "wait"}\\
			&\multicolumn{1}{c}{\textsc{Total}} &\multicolumn{1}{c}{\textsc{Correct}} &\multicolumn{1}{c}{\textsc{Incorrect}}&\multicolumn{1}{c}{\textsc{Total}} &\multicolumn{1}{c}{\textsc{Correct}} &\multicolumn{1}{c}{\textsc{Incorrect}}\\
			\midrule
			\textsc{DeepSeek-R1-Distill-Qwen}-1.5B &16390&8126&18775&107&49&129\\
			\textbf{\textsc{FastCuRL}-1.5B-Preview}  &8654&7091&12010&80&47&104\\
			\bottomrule
		\end{tabular}
	\end{center}
	\caption{Statistics of the outputs of \textsc{DeepSeek-R1-Distill-Qwen}-1.5B and \textbf{\textsc{FastCuRL}-1.5B-Preview}.}\label{length}
\end{table*}
\begin{figure*}[t!]
	\begin{center}
		\includegraphics[scale=0.317]{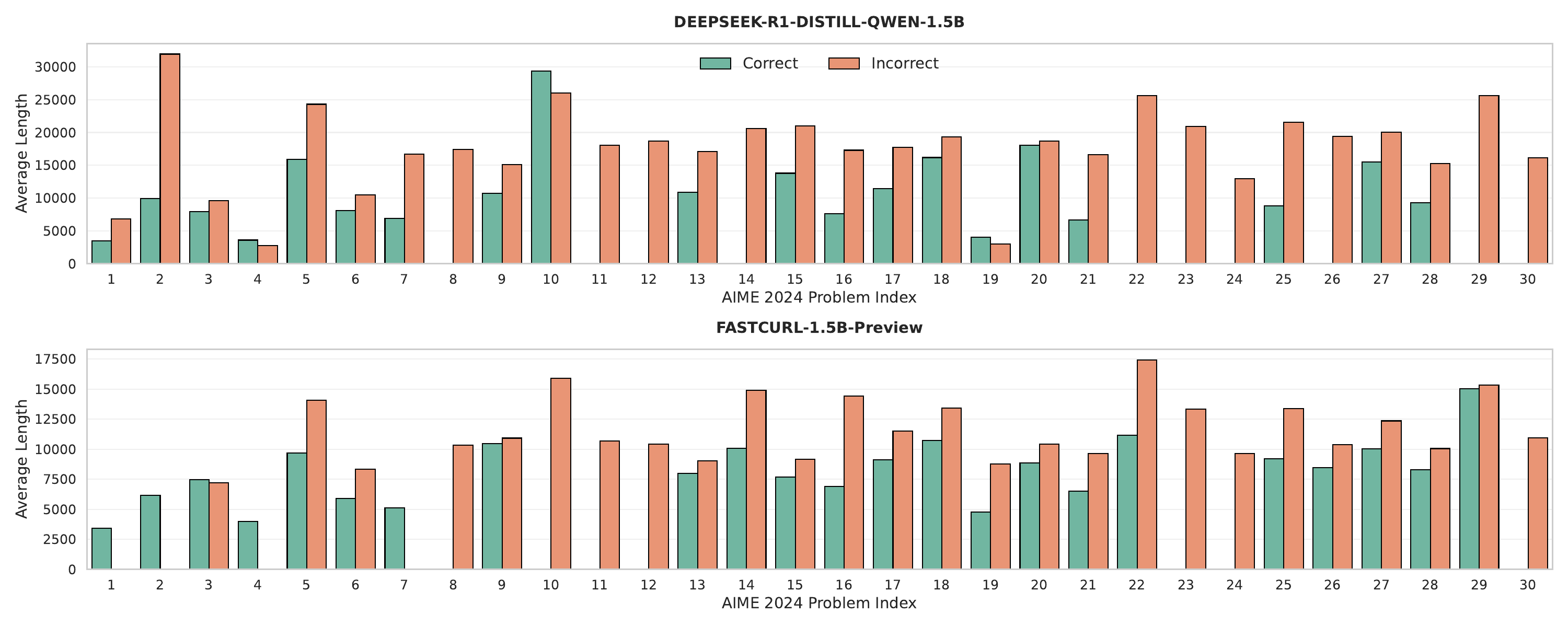}
	\end{center}
	\caption{Comparison of average response lengths (in tokens) between correct and incorrect answers for \textsc{DeepSeek-R1-Distill-Qwen}-1.5B (top) and \textsc{FastCuRL}-1.5B-Preview (bottom) on AIME 2024 problems. Green and orange bars represent correct and incorrect answers, respectively. The x-axis shows problem indices (1-30). The height of each bar reflects the average token count across 16 rollout solutions of each problem. Problems displaying only orange bars indicate cases where all 16 solutions failed to produce a correct answer.}\label{response_length}
\end{figure*}
\subsubsection{Dataset Complexity Verification}
To validate the effectiveness of our dataset selection approach, we train three reasoning models under identical settings on $\textbf{L}$1, $\textbf{L}$2, and $\textbf{L}$3 with an 8K context length. As illustrated in Figure~\ref{compare}, the experimental results demonstrate consistent patterns across all three metrics that align with our expectations. The output length progressively increases from $\textbf{L}$1 to $\textbf{L}$3 throughout the training process, with models trained on more complex datasets generating substantially longer responses, reflecting the additional reasoning steps required for solving complex problems. Correspondingly, the clip ratio exhibits the exact ordering ($\textbf{L}$3 > $\textbf{L}$2 > $\textbf{L}$1), indicating that complex problems naturally demand longer solution outputs that more frequently exceed the context length. While the reward scores show an inverse relationship, with $\textbf{L}$1 achieving the highest scores due to its simpler problems, this pattern is expected. This observation demonstrates the inherent trade-off between problem complexity and output length, which supports our hypothesis that problem complexity directly correlates with required output length; more complex problems necessitate longer, more detailed reasoning outputs to arrive at correct solutions, thereby validating the importance of complexity-aware data selection.
\subsubsection{Multi-Stage Experimental Results}
In our experiments, we conduct three sets of multi-stage experiments, comprising 3, 4, and 5 training stages, respectively. The specific parameter settings are detailed in Table 3, and the corresponding results are presented in Table~\ref{exp}.
In the first set of experiments, Exp-3 outperforms Exp-1 but requires additional training steps, as it is trained twice on the $\mathbf{L}$2 dataset. Considering the trade-off between performance and computational cost, we select Exp-1 as the base model for the second stage.
Following the first stage, we observe that the average response length is between 6,000 and 7,000 tokens. This observation motivates our subsequent evaluation of various context lengths. As shown in Table~\ref{exp}, a 16K context length (Exp-6) yields the best performance, surpassing the longer 24K and 32K settings. Consequently, we select Exp-6 as the base model for the third stage.

Informed by the second set of experiments, we conduct a third set evaluating context lengths of 24K, 16K, and 8K. As shown in Table~\ref{exp}, the 16K context still achieves the best performance, though it shows negligible improvement over the fourth stage. Our analysis of this phenomenon suggests that during compression and extension, the model's output length is initially constrained by the shorter context of the first stage. This constraint appears to compress the length of the thought process while enhancing its quality. 

As the context length extensions in subsequent stages, the model begins to explore problems requiring longer CoT outputs. However, this expansion improves the model's performance but also introduces repetitive thinking patterns. These patterns may not improve reasoning capabilities; on the contrary, they may diminish the model's exploratory efficiency, particularly at excessively long context lengths. Therefore, we posit that subsequent context compression, as performed in the fourth stage, is necessary to improve CoT quality and enhance exploratory efficiency.

\begin{figure}[t!]
	\begin{center}
		\includegraphics[scale=0.52]{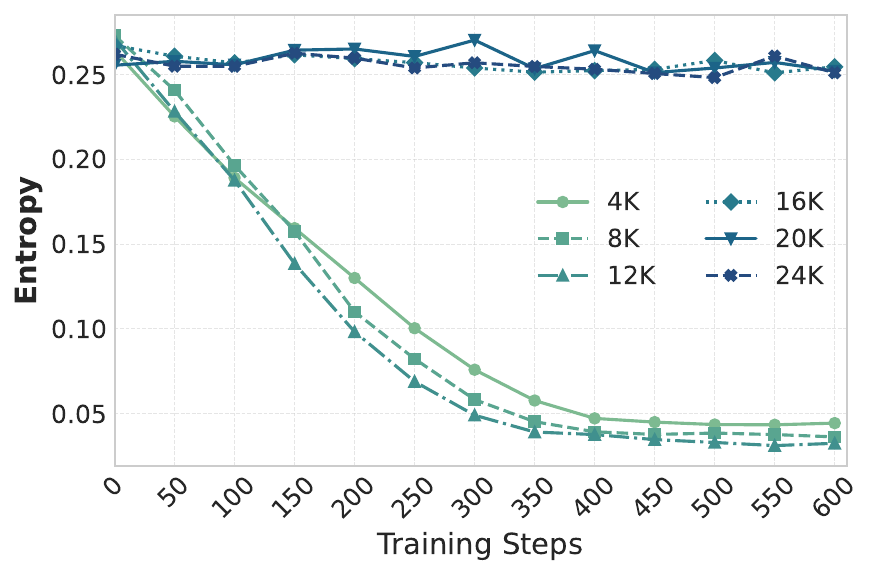}
	\end{center}
	\caption{Entropy curves of different context lengths. Here, data points are smoothed using a running average (window size = 10) and plotted at uniform intervals.}\label{entropy_loss}
\end{figure}

In the third set of experiments, we observe that neither increasing nor decreasing the context length from 16K yields comparable performance, suggesting that 16K represents an optimal configuration. This finding prompts a critical question: Does an optimal context length, or "sweet spot", exist for R1-like models? Specifically, we investigate whether 16K is the precise optimum for the \textsc{DeepSeek-R1-Distill-Qwen}-1.5B model or simply the best-performing value among the tested configurations of 8K and 24K.

To investigate this question, we conduct a comprehensive series of experiments examining model behavior across a finer-grained spectrum of context lengths. We train the model with various context length configurations while holding the entropy coefficient constant at $1\times10^{-6}$ to observe entropy dynamics. As illustrated in Figure~\ref{entropy_loss}, our findings reveal a notable dichotomy. For context lengths of 4K, 8K, and 12K, the entropy exhibits a rapid decline, indicating a significant degradation in the model's exploratory capabilities. In contrast, for context lengths of 16K, 20K, and 24K, the entropy demonstrates remarkable stability, converging to and maintaining a consistent value.

Inspired by the above findings, we continue to train \textbf{\textsc{FastCuRL}-1.5B-Preview} under a 16K context and adjust the coefficients of KL and Entropy (Table~\ref{exp}).
Results in Table~\ref{op} show that after being incentivized in the prior stages, the performance of \textbf{\textsc{FastCuRL}-1.5B-V3} gradually increases in Stage 5 and achieves an accuracy of 49.6\% on AIME 2024, supporting the above raised question.

\begin{figure}[t!]
	\begin{center}
		\includegraphics[scale=0.52]{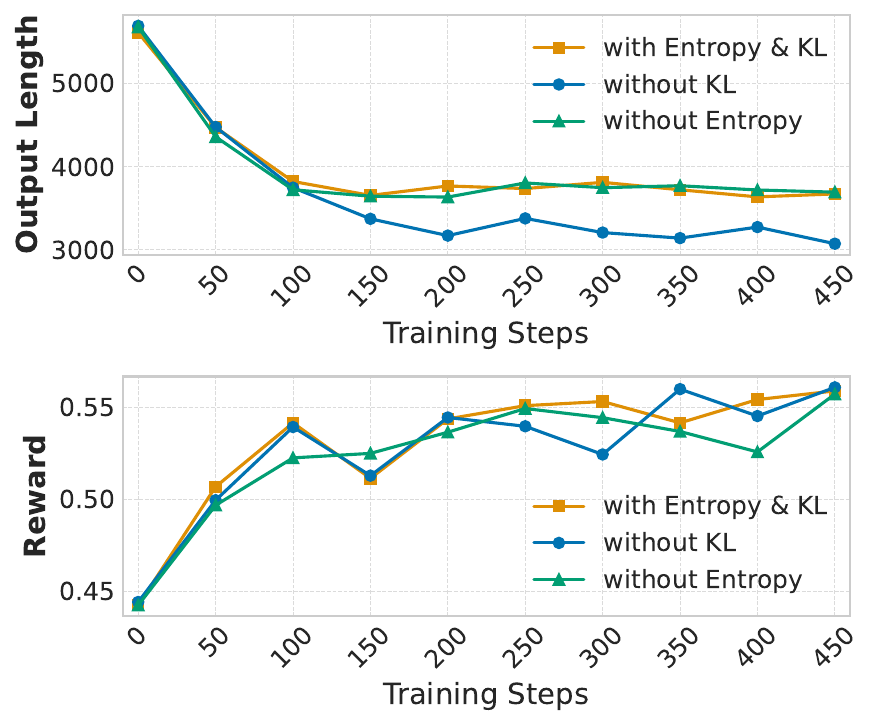}
	\end{center}
	\caption{Performance comparison of training with and without Entropy bonus at 8k context length.}\label{lr}
\end{figure}

\subsubsection{Analyzing Generated CoT Outputs}

Table~\ref{length} presents comparative statistics on the response characteristics of \textsc{DeepSeek-R1-Distill-Qwen}-1.5B and \textsc{FastCuRL}-1.5B-Preview. The results focus on two key metrics: average output length and frequency of the term "wait"/"Wait" in responses. The \textsc{DeepSeek-R1-Distill-Qwen}-1.5B produces significantly longer responses overall (50.5\% longer than \textsc{FastCuRL}-1.5B-Preview. Interestingly, both models show a pattern where incorrect responses tend to be substantially longer than correct ones. 
The frequency of "wait"/"Wait" terms is indicative of reflection behaviors in the R1-like reasoning models. \textsc{DeepSeek-R1-Distill-Qwen}-1.5B uses these terms approximately 36\% more frequently than \textsc{FastCuRL}-1.5B-Preview overall. Similarly, both models show significantly higher usage of these terms in incorrect responses compared to correct ones.

Figure~\ref{response_length} compares \textsc{FastCuRL}-1.5B-Preview and \textsc{DeepSeek-R1-Distill-Qwen}-1.5B on the AIME 2024, measuring the average response length between correct and incorrect answers at the problem level to observe and analyse whether the long incorrect response is related to the difficulty of the problem. Across both models, incorrect answers (red bars) almost universally have greater average response lengths than correct answers (green bars). This suggests that models tend to generate more verbose content when producing incorrect answers, potentially reflecting "over-explanation" or "verbose reasoning" when the model is uncertain.

\subsubsection{Ablation Study}
Figure~\ref{lr} presents a comprehensive comparison of model performance under different regularization configurations at an 8k context length. The upper panel demonstrates that models incorporating Entropy regularization (with Entropy \& KL) or KL divergence alone (without Entropy) maintain consistently higher output lengths around 3,700-3,900 tokens throughout training, while the model without KL divergence regularization exhibits notably shorter outputs, stabilizing at approximately 3,000-3,500 tokens. This suggests that KL divergence plays a critical role in preventing output length collapse during long-context generation. The lower panel reveals that all three configurations achieve comparable reward scores, converging to approximately 0.55 after initial training steps, indicating that the choice of regularization strategy primarily affects generation length rather than quality. Notably, the combination of entropy bonus and KL divergence (orange line) demonstrates the most stable training dynamics across both metrics, suggesting potential synergistic effects between these two regularization techniques in balancing exploration and exploitation during policy optimization.
\section{Related Work}
Advancements in RL methodologies have considerably enhanced the reasoning capabilities of LLMs. A pivotal development in this domain is OpenAI's o1 \cite{openai2024gpt4}, which employs RL training to promote the development of long CoT reasoning in LLMs. This approach has significantly enhanced performance on complex mathematical and programming benchmarks. Building upon this foundation, DeepSeek-R1 \cite{ds-r1} demonstrates that pure RL post-training via Group Reinforcement Policy Optimization (GRPO), without needing supervised pre-training, can directly perform robust CoT reasoning capabilities. Notably, this method not only achieves performance competitive with o1 but also exhibits emergent behaviors such as self-verification and multi-step planning. 
Building on these advancements, the research community has been collectively working to study and apply the methodology of DeepSeek-R1 to enhance the CoT reasoning capabilities of various sizes of language models, yielding remarkable progress, such as~\citet{openr1,deepscaler2025, zeng2025simplerl,drgrpo,dapo, song2025walkrunconcisellm}.
Different from the existing approaches, we employ an iterative stage-wise scaling method that alternates between CoT compression (long-to-short) and CoT extension (short-to-long) training stages. We initially compress the CoT outputs, subsequently perform extension, followed by re-compression and re-extension in a cyclical manner. Specifically, we hypothesize that this dynamic interplay between compression and extension facilitates the emergence of higher-quality reasoning responses by iteratively refining the essential logical dependencies while eliminating redundant or spurious CoT reasoning outputs.
\section{Conclusion}
We investigate how the model's context length and the complexity of the training dataset influence the training process of R1-like reasoning models. Motivated by our findings, we propose \textsc{\textbf{FastCuRL}}, a simple yet effective curriculum reinforcement learning framework incorporating a stage-wise context scaling strategy. This method is designed to accelerate the training efficiency and improve the model's long CoT reasoning capabilities. Experimental results demonstrate that \textsc{FastCuRL}-1.5B-Preview achieves better performance and reduces computational resource consumption by more than 50\%, with all training phases efficiently executed using a single node with 8 GPUs.

\section{Limitations}
In this paper, we verify the proposed methodology exclusively on a 1.5B parameter language model. Extending the proposed methodology to models of diverse scales represents a critical avenue for future investigation. Moreover, while this work demonstrates the substantial impact of complexity-aware data selection through a relatively straightforward stratification approach, the observed performance gains suggest that more sophisticated data selection methodologies could potentially yield superior outcomes. 
Meanwhile, this observation motivates several research directions, including the development of adaptive context length and dynamic KL penalty controlling mechanisms, both of which could further optimize the balance between exploration and convergence in preference learning.
\bibliography{custom}

\begin{thebibliography}{29}
\providecommand{\natexlab}[1]{#1}

\bibitem[{Balunović et~al.(2025)Balunović, Dekoninck, Petrov, Jovanović, and
  Vechev}]{matharena}
Mislav Balunović, Jasper Dekoninck, Ivo Petrov, Nikola Jovanović, and Martin
  Vechev. 2025.
\newblock \href {https://matharena.ai/} {Matharena: Evaluating llms on
  uncontaminated math competitions}.

\bibitem[{Chen et~al.(2025)Chen, Qin, Liu, Peng, Guan, Wang, Hu, Zhou, Gao, and
  Che}]{chen2025reasoningerasurveylong}
Qiguang Chen, Libo Qin, Jinhao Liu, Dengyun Peng, Jiannan Guan, Peng Wang,
  Mengkang Hu, Yuhang Zhou, Te~Gao, and Wanxiang Che. 2025.
\newblock \href {https://arxiv.org/abs/2503.09567} {Towards reasoning era: A
  survey of long chain-of-thought for reasoning large language models}.
\newblock \emph{Preprint}, arXiv:2503.09567.

\bibitem[{Cui et~al.(2025)Cui, Yuan, Wang, Wang, Li, He, Fan, Yu, Xu, Chen,
  Yuan, Chen, Zhang, Lv, Wang, Yao, Han, Peng, Cheng, Liu, Sun, Zhou, and
  Ding}]{Eurus-2-7B-PRIME}
Ganqu Cui, Lifan Yuan, Zefan Wang, Hanbin Wang, Wendi Li, Bingxiang He, Yuchen
  Fan, Tianyu Yu, Qixin Xu, Weize Chen, Jiarui Yuan, Huayu Chen, Kaiyan Zhang,
  Xingtai Lv, Shuo Wang, Yuan Yao, Xu~Han, Hao Peng, Yu~Cheng, and 4 others.
  2025.
\newblock \href {https://arxiv.org/abs/2502.01456} {Process reinforcement
  through implicit rewards}.
\newblock \emph{Preprint}, arXiv:2502.01456.

\bibitem[{DeepSeek-AI(2025)}]{ds-r1}
DeepSeek-AI. 2025.
\newblock \href {https://arxiv.org/abs/2501.12948} {Deepseek-r1: Incentivizing
  reasoning capability in llms via reinforcement learning}.
\newblock \emph{Preprint}, arXiv:2501.12948.

\bibitem[{Face(2025)}]{openr1}
Hugging Face. 2025.
\newblock \href {https://github.com/huggingface/open-r1} {Open r1: A fully open
  reproduction of deepseek-r1}.

\bibitem[{Gao et~al.(2024)Gao, Song, Yang, Cai, Miao, Dong, Li, Ma, Chen, Xu,
  Tang, Wang, Zan, Quan, Zhang, Sha, Zhang, Ren, Liu, and Chang}]{OmniMATH}
Bofei Gao, Feifan Song, Zhe Yang, Zefan Cai, Yibo Miao, Qingxiu Dong, Lei Li,
  Chenghao Ma, Liang Chen, Runxin Xu, Zhengyang Tang, Benyou Wang, Daoguang
  Zan, Shanghaoran Quan, Ge~Zhang, Lei Sha, Yichang Zhang, Xuancheng Ren,
  Tianyu Liu, and Baobao Chang. 2024.
\newblock Omni-math: {A} universal olympiad level mathematic benchmark for
  large language models.
\newblock \emph{CoRR}, abs/2410.07985.

\bibitem[{Guan et~al.(2025)Guan, Zhang, Liu, Shang, Sun, Zhu, Yang, and
  Yang}]{rStar-Math}
Xinyu Guan, Li~Lyna Zhang, Yifei Liu, Ning Shang, Youran Sun, Yi~Zhu, Fan Yang,
  and Mao Yang. 2025.
\newblock \href {https://arxiv.org/abs/2501.04519} {rstar-math: Small llms can
  master math reasoning with self-evolved deep thinking}.
\newblock \emph{Preprint}, arXiv:2501.04519.

\bibitem[{He et~al.(2024)He, Luo, Bai, Hu, Thai, Shen, Hu, Han, Huang, Zhang,
  Liu, Qi, Liu, and Sun}]{OlympiadBench}
Chaoqun He, Renjie Luo, Yuzhuo Bai, Shengding Hu, Zhen~Leng Thai, Junhao Shen,
  Jinyi Hu, Xu~Han, Yujie Huang, Yuxiang Zhang, Jie Liu, Lei Qi, Zhiyuan Liu,
  and Maosong Sun. 2024.
\newblock Olympiadbench: {A} challenging benchmark for promoting {AGI} with
  olympiad-level bilingual multimodal scientific problems.
\newblock In \emph{{ACL} {(1)}}, pages 3828--3850. Association for
  Computational Linguistics.

\bibitem[{Hendrycks et~al.(2021)Hendrycks, Burns, Kadavath, Arora, Basart,
  Tang, Song, and Steinhardt}]{MATH}
Dan Hendrycks, Collin Burns, Saurav Kadavath, Akul Arora, Steven Basart, Eric
  Tang, Dawn Song, and Jacob Steinhardt. 2021.
\newblock \href
  {https://datasets-benchmarks-proceedings.neurips.cc/paper/2021/hash/be83ab3ecd0db773eb2dc1b0a17836a1-Abstract-round2.html}
  {Measuring mathematical problem solving with the {MATH} dataset}.
\newblock In \emph{Proceedings of the Neural Information Processing Systems
  Track on Datasets and Benchmarks 1, NeurIPS Datasets and Benchmarks 2021,
  December 2021, virtual}.

\bibitem[{Hu et~al.(2025)Hu, Zhang, Han, Jiang, and
  Xiangyu~Zhang}]{OpenReasonerZero2025}
Jingcheng Hu, Yinmin Zhang, Qi~Han, Daxin Jiang, and Heung-Yeung~Shum
  Xiangyu~Zhang. 2025.
\newblock Open-reasoner-zero: An open source approach to scaling reinforcement
  learning on the base model.
\newblock \url{https://github.com/Open-Reasoner-Zero/Open-Reasoner-Zero}.

\bibitem[{Lewkowycz et~al.(2022)Lewkowycz, Andreassen, Dohan, Dyer,
  Michalewski, Ramasesh, Slone, Anil, Schlag, Gutman-Solo, Wu, Neyshabur,
  Gur-Ari, and Misra}]{Minerva}
Aitor Lewkowycz, Anders Andreassen, David Dohan, Ethan Dyer, Henryk
  Michalewski, Vinay Ramasesh, Ambrose Slone, Cem Anil, Imanol Schlag, Theo
  Gutman-Solo, Yuhuai Wu, Behnam Neyshabur, Guy Gur-Ari, and Vedant Misra.
  2022.
\newblock \href {https://arxiv.org/abs/2206.14858} {Solving quantitative
  reasoning problems with language models}.
\newblock \emph{Preprint}, arXiv:2206.14858.

\bibitem[{Liu et~al.(2025)Liu, Chen, Li, Qi, Pang, Du, Lee, and Lin}]{drgrpo}
Zichen Liu, Changyu Chen, Wenjun Li, Penghui Qi, Tianyu Pang, Chao Du, Wee~Sun
  Lee, and Min Lin. 2025.
\newblock Understanding r1-zero-like training: A critical perspective.
\newblock \emph{arXiv preprint arXiv:2503.20783}.

\bibitem[{Luo et~al.(2025)Luo, Tan, Wong, Shi, Tang, Roongta, Cai, Luo, Zhang,
  Li, Popa, and Stoica}]{deepscaler2025}
Michael Luo, Sijun Tan, Justin Wong, Xiaoxiang Shi, William~Y. Tang, Manan
  Roongta, Colin Cai, Jeffrey Luo, Tianjun Zhang, Li~Erran Li, Raluca~Ada Popa,
  and Ion Stoica. 2025.
\newblock Deepscaler: Surpassing o1-preview with a 1.5b model by scaling rl.
\newblock \url{https://github.com/agentica-project/deepscaler}.
\newblock Notion Blog.

\bibitem[{Min et~al.(2024)Min, Chen, Jiang, Chen, Deng, Hu, Tang, Wang, Cheng,
  Song, Zhao, Liu, Wang, and Wen}]{still_dataset}
Yingqian Min, Zhipeng Chen, Jinhao Jiang, Jie Chen, Jia Deng, Yiwen Hu, Yiru
  Tang, Jiapeng Wang, Xiaoxue Cheng, Huatong Song, Wayne~Xin Zhao, Zheng Liu,
  Zhongyuan Wang, and Ji-Rong Wen. 2024.
\newblock \href {https://arxiv.org/abs/2412.09413} {Imitate, explore, and
  self-improve: A reproduction report on slow-thinking reasoning systems}.
\newblock \emph{Preprint}, arXiv:2412.09413.

\bibitem[{Minaee et~al.(2024)Minaee, Mikolov, Nikzad, Chenaghlu, Socher,
  Amatriain, and Gao}]{llm_survey_2024}
Shervin Minaee, Tomas Mikolov, Narjes Nikzad, Meysam Chenaghlu, Richard Socher,
  Xavier Amatriain, and Jianfeng Gao. 2024.
\newblock \href {https://arxiv.org/abs/2402.06196} {Large language models: A
  survey}.
\newblock \emph{Preprint}, arXiv:2402.06196.

\bibitem[{Muennighoff et~al.(2025)Muennighoff, Yang, Shi, Li, Fei-Fei,
  Hajishirzi, Zettlemoyer, Liang, Cand{\`e}s, and
  Hashimoto}]{muennighoff2025s1}
Niklas Muennighoff, Zitong Yang, Weijia Shi, Xiang~Lisa Li, Li~Fei-Fei,
  Hannaneh Hajishirzi, Luke Zettlemoyer, Percy Liang, Emmanuel Cand{\`e}s, and
  Tatsunori Hashimoto. 2025.
\newblock s1: Simple test-time scaling.
\newblock \emph{arXiv preprint arXiv:2501.19393}.

\bibitem[{OpenAI(2024)}]{openai2024gpt4}
OpenAI. 2024.
\newblock \href {https://arxiv.org/abs/2303.08774} {Gpt-4 technical report}.
\newblock \emph{Preprint}, arXiv:2303.08774.

\bibitem[{Shao et~al.(2024)Shao, Wang, Zhu, Xu, Song, Bi, Zhang, Zhang, Li, Wu,
  and Guo}]{deepseek-math}
Zhihong Shao, Peiyi Wang, Qihao Zhu, Runxin Xu, Junxiao Song, Xiao Bi, Haowei
  Zhang, Mingchuan Zhang, Y.~K. Li, Y.~Wu, and Daya Guo. 2024.
\newblock \href {https://arxiv.org/abs/2402.03300} {Deepseekmath: Pushing the
  limits of mathematical reasoning in open language models}.
\newblock \emph{Preprint}, arXiv:2402.03300.

\bibitem[{Snell et~al.(2024)Snell, Lee, Xu, and Kumar}]{tts}
Charlie Snell, Jaehoon Lee, Kelvin Xu, and Aviral Kumar. 2024.
\newblock \href {https://arxiv.org/abs/2408.03314} {Scaling llm test-time
  compute optimally can be more effective than scaling model parameters}.
\newblock \emph{Preprint}, arXiv:2408.03314.

\bibitem[{Song and Zheng(2025)}]{song2025walkrunconcisellm}
Mingyang Song and Mao Zheng. 2025.
\newblock \href {https://arxiv.org/abs/2505.21178} {Walk before you run!
  concise llm reasoning via reinforcement learning}.
\newblock \emph{Preprint}, arXiv:2505.21178.

\bibitem[{Team(2025{\natexlab{a}})}]{kimi1.5}
Kimi Team. 2025{\natexlab{a}}.
\newblock \href {https://arxiv.org/abs/2501.12599} {Kimi k1.5: Scaling
  reinforcement learning with llms}.
\newblock \emph{Preprint}, arXiv:2501.12599.

\bibitem[{Team(2025{\natexlab{b}})}]{Slow_Thinking_with_LLMs_3_Preview}
RUCAIBox~STILL Team. 2025{\natexlab{b}}.
\newblock \href {https://github.com/RUCAIBox/Slow_Thinking_with_LLMs}
  {Still-3-1.5b-preview: Enhancing slow thinking abilities of small models
  through reinforcement learning}.

\bibitem[{Wei et~al.(2023)Wei, Wang, Schuurmans, Bosma, Ichter, Xia, Chi, Le,
  and Zhou}]{cot}
Jason Wei, Xuezhi Wang, Dale Schuurmans, Maarten Bosma, Brian Ichter, Fei Xia,
  Ed~Chi, Quoc Le, and Denny Zhou. 2023.
\newblock \href {https://arxiv.org/abs/2201.11903} {Chain-of-thought prompting
  elicits reasoning in large language models}.
\newblock \emph{Preprint}, arXiv:2201.11903.

\bibitem[{Wu et~al.(2025)Wu, Wang, Du, Jegelka, and
  Wang}]{wu2025lessunderstandingchainofthoughtlength}
Yuyang Wu, Yifei Wang, Tianqi Du, Stefanie Jegelka, and Yisen Wang. 2025.
\newblock \href {https://arxiv.org/abs/2502.07266} {When more is less:
  Understanding chain-of-thought length in llms}.
\newblock \emph{Preprint}, arXiv:2502.07266.

\bibitem[{Yang et~al.(2024)Yang, Zhang, Hui, Gao, Yu, Li, Liu, Tu, Zhou, Lin,
  Lu, Xue, Lin, Liu, Ren, and Zhang}]{Qwen2.5-Math}
An~Yang, Beichen Zhang, Binyuan Hui, Bofei Gao, Bowen Yu, Chengpeng Li,
  Dayiheng Liu, Jianhong Tu, Jingren Zhou, Junyang Lin, Keming Lu, Mingfeng
  Xue, Runji Lin, Tianyu Liu, Xingzhang Ren, and Zhenru Zhang. 2024.
\newblock \href {https://doi.org/10.48550/ARXIV.2409.12122} {Qwen2.5-math
  technical report: Toward mathematical expert model via self-improvement}.
\newblock \emph{CoRR}, abs/2409.12122.

\bibitem[{Yeo et~al.(2025)Yeo, Tong, Niu, Neubig, and
  Yue}]{yeo2025demystifyinglongchainofthoughtreasoning}
Edward Yeo, Yuxuan Tong, Morry Niu, Graham Neubig, and Xiang Yue. 2025.
\newblock \href {https://arxiv.org/abs/2502.03373} {Demystifying long
  chain-of-thought reasoning in llms}.
\newblock \emph{Preprint}, arXiv:2502.03373.

\bibitem[{Yu et~al.(2025)Yu, Zhang, Zhu, Yuan, Zuo, Yue, Fan, Liu, Liu, Liu,
  Lin, Lin, Ma, Sheng, Tong, Zhang, Zhang, Zhang, Zhu, Zhu, Chen, Chen, Wang,
  Yu, Dai, Song, Wei, Zhou, Liu, Ma, Zhang, Yan, Qiao, Wu, and Wang}]{dapo}
Qiying Yu, Zheng Zhang, Ruofei Zhu, Yufeng Yuan, Xiaochen Zuo, Yu~Yue, Tiantian
  Fan, Gaohong Liu, Lingjun Liu, Xin Liu, Haibin Lin, Zhiqi Lin, Bole Ma,
  Guangming Sheng, Yuxuan Tong, Chi Zhang, Mofan Zhang, Wang Zhang, Hang Zhu,
  and 16 others. 2025.
\newblock \href {https://arxiv.org/abs/2503.14476} {Dapo: An open-source llm
  reinforcement learning system at scale}.
\newblock \emph{Preprint}, arXiv:2503.14476.

\bibitem[{Zeng et~al.(2025)Zeng, Huang, Liu, He, Liu, Ma, and
  He}]{zeng2025simplerl}
Weihao Zeng, Yuzhen Huang, Wei Liu, Keqing He, Qian Liu, Zejun Ma, and Junxian
  He. 2025.
\newblock 7b model and 8k examples: Emerging reasoning with reinforcement
  learning is both effective and efficient.
\newblock \url{https://hkust-nlp.notion.site/simplerl-reason}.
\newblock Notion Blog.

\bibitem[{Zhao et~al.(2023)Zhao, Zhou, Li, Tang, Wang, Hou, Min, Zhang, Zhang,
  Dong, Du, Yang, Chen, Chen, Jiang, Ren, Li, Tang, Liu, Liu, Nie, and
  Wen}]{llm_survey_2023}
Wayne~Xin Zhao, Kun Zhou, Junyi Li, Tianyi Tang, Xiaolei Wang, Yupeng Hou,
  Yingqian Min, Beichen Zhang, Junjie Zhang, Zican Dong, Yifan Du, Chen Yang,
  Yushuo Chen, Zhipeng Chen, Jinhao Jiang, Ruiyang Ren, Yifan Li, Xinyu Tang,
  Zikang Liu, and 3 others. 2023.
\newblock \href {https://arxiv.org/abs/2303.18223} {A survey of large language
  models}.
\newblock \emph{Preprint}, arXiv:2303.18223.

\end{thebibliography}
\end{document}